\documentclass[letterpaper, 10 pt, conference]{ieeeconf}  

\IEEEoverridecommandlockouts                              

\usepackage[utf8]{inputenc}

\usepackage[l2tabu,orthodox]{nag}

\usepackage[
backend=biber, 
style=ieee, 
sorting=none, 
natbib=true, 
doi=false, 
isbn=false, 
url=false, 
eprint=false, 
maxcitenames=1, 
mincitenames=1
]{biblatex}

\usepackage[pdftex,colorlinks]{hyperref}

\usepackage[printonlyused]{acronym}

\usepackage{siunitx}
\usepackage[all]{nowidow}

\usepackage[dvipsnames]{xcolor}

\usepackage{lipsum}

\usepackage[pdftex]{graphicx}

\usepackage{subcaption}

\usepackage{epstopdf}

\usepackage{import}

\graphicspath{{./figures/}}

\usepackage{booktabs}

\usepackage{array}
\usepackage{tabularx}
\usepackage{multirow}
\usepackage{threeparttable}

\usepackage{amssymb,amsfonts,amsmath,amscd}

\usepackage{bm} 

\newcommand{\norm}[1]{\left\Vert#1\right\Vert}
\newcommand{\abs}[1]{\left\vert#1\right\vert}

\newcommand{\bbm}{\begin{bmatrix}}
\newcommand{\ebm}{\end{bmatrix}}
\newcommand{\diff}{\mathop{}\!\mathrm{d}}

\newcommand{\T}{^{T}}

\title{\LARGE \bf Geometry Preserving Sampling Method based on Spectral Decomposition for 3D Registration}%
\author{Mathieu Labussiere\authorrefmark{1}\authorrefmark{2}, Johann Laconte\authorrefmark{1}\authorrefmark{2} and Fran\c{c}ois Pomerleau\authorrefmark{2}%
\thanks{\authorrefmark{1}The authors are with Institut Pascal, Campus des Cezeaux, 63178 Aubière Cedex, France. 
	{\tt\small mathieu.labu@gmail.com}}%
\thanks{\authorrefmark{2}The authors are with Northern Robotics Laboratory, Universit\'{e} Laval, Canada.
	{\tt\small francois.pomerleau@ift.ulaval.ca}}%
}

\acrodef{ICP}{Iterative Closest Point}
\acrodef{RANSAC}{Random Sample Consensus}
\acrodef{LIDAR}{Light Detection And Ranging}
\acrodef{L2}{squared distance}
\acrodef{MLS}{Moving Least-Squares}

\acrodef{FPFH}{Fast Point Feature Histogram}

\acrodef{NSS}{Normal-Space Sampling}
\acrodef{DNSS}{Dual Normal-Space Sampling}
\acrodef{CovS}{Covariance Sampling}

\acrodef{SpDF}{Spectral Decomposition Filter}
\acrodef{CFTV}{closed-form}
\acrodef{TV}{Tensor Voting}

\begin{document}

\maketitle
\thispagestyle{empty}
\pagestyle{empty}

\begin{abstract}
In the context of 3D mapping, larger and larger point clouds are acquired with LIDAR sensors.
The Iterative Closest Point (ICP) algorithm is used to align these point clouds.
However, its complexity is directly dependent of the number of points to process. 
Several strategies exist to address this problem by reducing the number of points.
However, they tend to underperform with non-uniform density, large sensor noise, spurious measurements, and large-scale point clouds, which is the case in mobile robotics.
This paper presents a novel sampling algorithm for registration in ICP algorithm based on spectral decomposition analysis and called Spectral Decomposition Filter (SpDF).
It preserves geometric information along the topology of point clouds and is able to scale to large environments with non-uniform density.
The effectiveness of our method is validated and illustrated by quantitative and qualitative experiments on various environments.%
\end{abstract}


\section{Introduction}
\ac{LIDAR} sensors has recently been democratized in robotics applications.
These sensors are able to acquire an efficient representation of the environment (i.e., a point cloud), which can be used in localization algorithms or for 3D mapping.

Such algorithms rely on point cloud registration. 
Registration is the process of aligning the frames of two point clouds, the reference $\bm{P}$ and the reading $\bm{Q}$, by finding the rigid transformation $\bm{T} \in \mathrm{SE}(3)$ between them by a minimization process.
The transformation can be determined through the \ac{ICP} algorithm introduced by~\citet{Besl1992a,Chen1992}, and still considered a strong solution for registration in mobile robotics~\cite{Pomerleau2015b}. 

\begin{figure}[t]
		\centering
		\vspace{-8pt}
		\includegraphics[width=0.90\columnwidth]{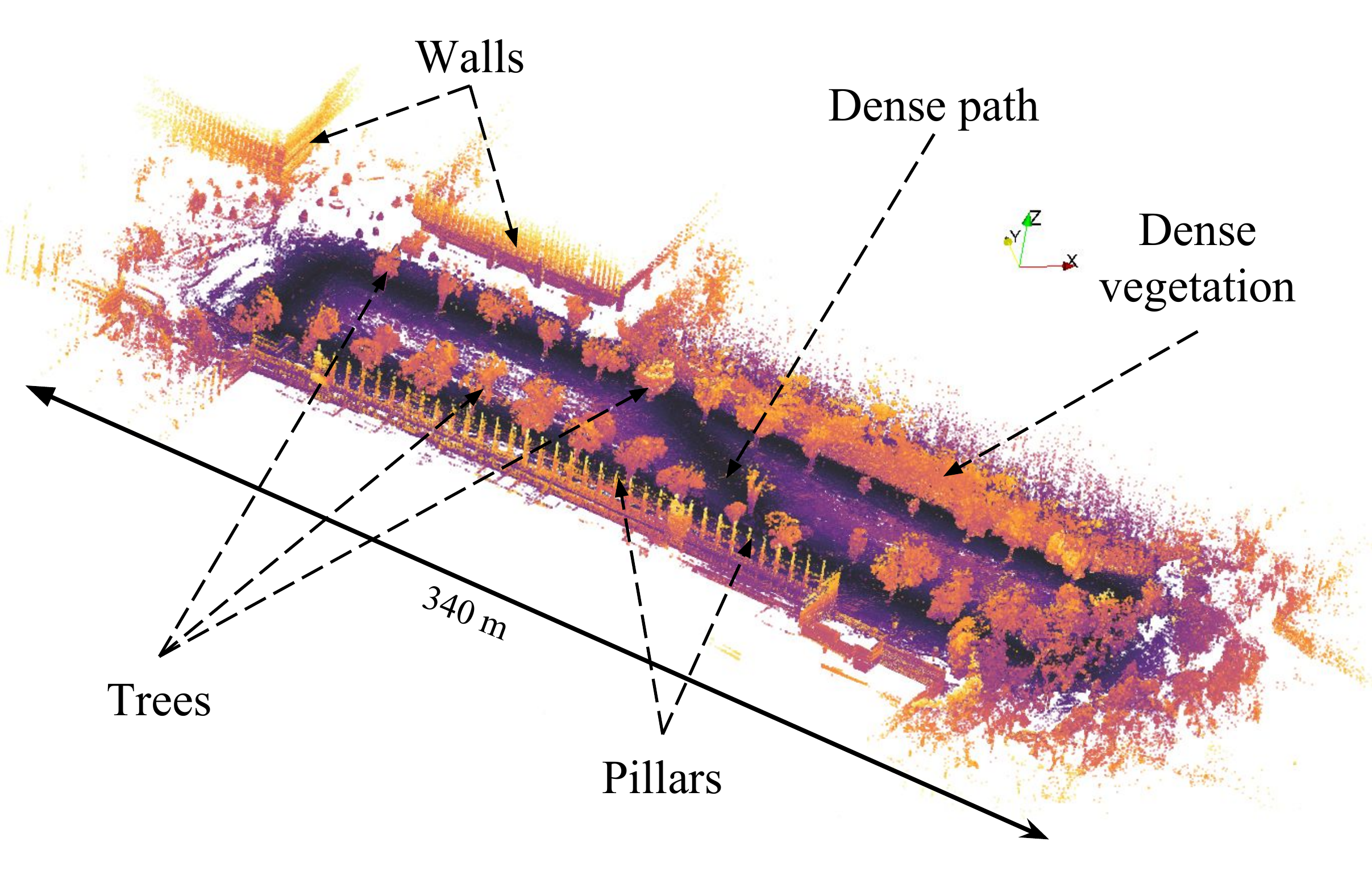}%
		\vspace*{-8pt}
		\hspace*{-15pt}\includegraphics[width=0.60\columnwidth]{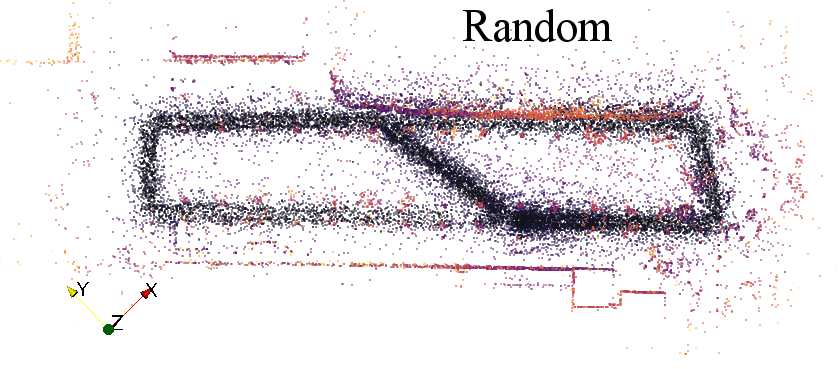}%
		\hspace*{-15pt}\includegraphics[width=0.60\columnwidth]{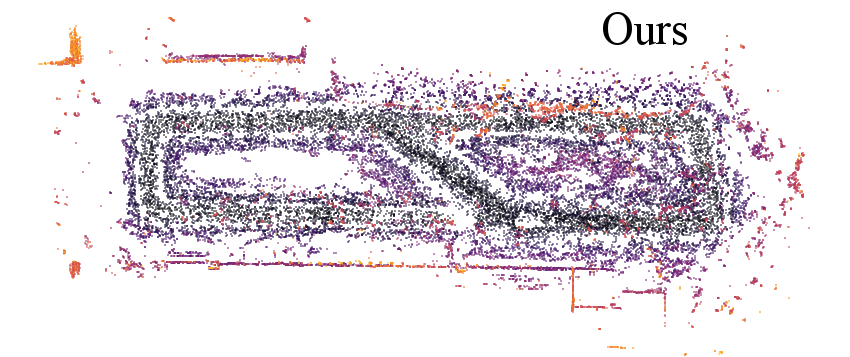}
		\vspace*{-15pt}
		\caption{
			This large-scale map, containing more than \num{4.6} million points, presents structured (walls, pillars) and unstructured (trees, vegetation) elements with varying densities.
			Brighter the color, farther the point has been observed.
			\emph{Bottom}: Top view of the reduced map to \num{30}\si{\kilo} points by random sampling (\emph{Left}) where mostly dense areas have been kept, losing lots of details; 
			by our proposed method (\emph{Right}) where the density is uniform and more geometric details have been preserved. 
		}
		\label{fig:laval}
		\vspace{-15pt}
	\end{figure}

Prior works on \ac{LIDAR}-based registration algorithms have been recently used to create larger and larger 3D maps~\cite{Pomerleau2015}.
As an example, \autoref{fig:laval} shows the map of the ``Grand Axe'' of Laval University campus, where only a few hours of data collection lead to a number of points at the limit of real-time computation capability. 
The sensor used for this map, the Velodyne HDL-32E, yields up to \num{1.39} million points per second.
The problem of limiting the growth of large point clouds during mapping is typically addressed by using an efficient representation based on octrees with a compression scheme to reduce the amount of data at a similar location~\cite{Elseberg2013,Lalonde:2007ib}.
Although pleasing to the eye, dense maps are not necessarily tailored for accurate registration.%
%
The complexity of the \ac{ICP} algorithm (i.e., the computation time) is directly dependent on the number of input points~\cite{Pomerleau2015b}.
On one hand, limiting the growth by reducing their number will enlarge the spectrum of real-time mapping applications.
On the other hand, subsampling the points too aggressively can lead to less accurate localization.
Primary investigations on the influence of the number of points on registration accuracy show that preserving and taking into account topology (i.e., the spatial distribution) leads to less errors in translation and rotation. 
Furthermore, geometric primitives are able to capture the details along the topology and must be preserved as much as possible.
These information can be retrieved for instance by the methodology of Tensor Voting~\cite{Guy1997,Medioni2000}. 

In the context of mobile robotics, data are typically noisy and sparse due to their radial distribution around \ac{LIDAR} sensors.
As opposed to solutions for registration-based object reconstruction, we will consider large-scale 3D maps, which are challenging even for the state-of-the-art sampling methods presented in \autoref{sec:relatedwork}.
Given these working hypotheses, the contributions of this paper can be summarized as follows:
\begin{enumerate}
	\item A novel subsampling method, called \ac{SpDF} and based on spectral decomposition analysis, preserving geometric information along the topology of point clouds, and able to scale to large environments.
	\item A solution to tensor voting limitations with uneven distributions of points in large-scale 3D maps, by proposing a new procedure to ensure uniformity on each geometric primitive.
	\item A large-scale comparison of current subsampling strategies relying on \num{2.95} million registrations in different types of environments.
\end{enumerate}

\section{Related Works}\label{sec:relatedwork}

A point-sampled surface is a good representation for analyzing the properties of 3D shapes~\cite{Alexa2001}. 
Unfortunately, most point clouds obtained in robotics context are noisy, sparse, large and have an uneven density.
%
An important step during the process of analyzing point clouds is to remove noise and outliers. 
This can be done using filtering algorithms.
An extensive review of these algorithms has been realized by~\citet{Han2017}.
Point cloud simplification is related to the problematic addressed by the computer vision field but aims to accelerate graphic rendering.
A lot of methods based on meshing are used to address this problem. 
Technically, these methods can be directly extended to point cloud representation, but most of the algorithms perform an expensive dataset meshing pre-step. 
A review and comparison of mesh simplification algorithms has been done by~\citet{Cignoni1998}. 
Mesh-free algorithms have also been developed to directly simplify point clouds. 
For instance,~\citet{Pauly2002} introduced, analyzed and quantitatively compared a number of surface simplification methods for point-sampled geometry. 
Most of these algorithms however cannot be used with noisy and sparse point clouds as they are rarely designed for robotics applications.

Sampling algorithms aim to decrease the complexity of \ac{ICP} (i.e., the computation time) by  reducing the number of input points. 
There are different strategies for points selection that can be categorized as global methods (e.g., uniform and random sampling, spatial sampling), local methods (e.g., using geometric information), and feature-based methods.
Feature-based methods such as \ac{FPFH} introduced by~\citet{Rusu2009} use features which describe the local geometry around a point.
It reduces then the number of points by grouping them to describe the neighborhood.
These methods provide improvements only with point clouds where features are distinctive which is hard to obtain with noise or incomplete data~\cite{Mellado2014}.
Hence, this paper will only focus on global and local methods. 

Another category of methods analyzes geometric primitives to sample relevant points in point clouds. 
For instance, the curvature provides a lot of information (e.g.,~\citet{Rodola2015} defined the concept of relevance based on curvature to sample points) but such primitives are often noisy and must be processed carefully.
\citet{Rusinkiewicz2001} proposed a method based on normals analysis named \ac{NSS}. 
It helps convergence for scenes with small, sparse features but cannot handle rotational uncertainties. 
\citet{Kwok2018} extended the latter work to handle them by introducing a dual normal space to constrain both translation and rotation. 
Both \ac{NSS} and \ac{DNSS} do not take into account the spatial distribution of the sampling points, leading to less accurate results in large-scale sparse point clouds.
\citet{Gelfand2003} presented a method based on covariance analysis, \ac{CovS}, to perform stability analysis in order to select geometrically stable points that can bind the rotational component as well as the translation. 
It is efficient when the algorithm is able to detect constraining areas but the impact is negligible otherwise. 
An improvement of \ac{CovS} has been proposed in the context of manufacturing~\cite{Kwok2015}. 
Given our results, previous methods cannot handle large-scale and density-varying point clouds when they are used to reduce the number of points.
%
Some methods based on octree or voxel representations of point clouds~\cite{Elseberg2013,hornung2013octomap} can take into consideration the spatial distribution of the points. 
We can then reduce the number of points by taking the most representative point in each cell, e.g., the centroid. 
Spatial segmentation methods however do not take into account distinctiveness of geometric features, losing important information in dense areas as our results will show.

Eventually, the sampling process can be addressed by signal processing strategies. 
Indeed, a point cloud can be considered as a manifold sample. 
\citet{Pauly2001} introduced the concept of local frequencies on geometry in order to be able to use all existing signal processing algorithms. 
\citet{Oztireli2010} proposed a new method to find optimal sampling conditions based on spectral analysis of manifolds. 
However, these methods stand under the hypothesis of smooth manifolds, which is rarely the case in maps acquired with \ac{LIDAR} sensors in robotics.
%
The method presented in this paper is able to both retrieve the geometric information in large sparse noisy point clouds and take into account the spatial distribution of the points.

\section{Tensor Voting: Theory}\label{sec:tvtheory}

\citet{Medioni2000} introduced \ac{TV} as a methodology to infer geometric information (e.g., surface, curve, and junction descriptions) from sparse 3D data\footnote{The original formulation can be extended to \textit{n}-D data~\cite{Tang2001}.}. 
The algorithm is based on tensor calculus for data representation and tensor voting for data communication.
Theory related to \ac{TV} will be summarized in this section for completeness.

	\subsubsection{Tensor Representation}
	%
	To capture the first order differential geometry information and its saliency, each datum can be represented as a second order symmetric tensor in the normal space. 
	In 3D, such a tensor can be visualized as an ellipsoid with a shape that defines the nature of the information and a scale that defines the saliency of this information.
	A second order symmetric tensor $\bm{K}$ is fully described by its associated spectral decomposition using three eigenvectors $\bm{e}_1$, $\bm{e}_2$ and $\bm{e}_3$, and three corresponding ordered positive eigenvalues $\lambda_1 \ge \lambda_2 \ge \lambda_3$. 
	This tensor can be decomposed in three basis tensors, resulting in
	\begin{equation}
		\bm{K} = \left(\lambda_1 - \lambda_2 \right) \bm{S} + \left(\lambda_2 - \lambda_3 \right) \bm{P} + \lambda_3 \bm{B},
	\end{equation}
	with 
	\begin{equation}
	\begin{aligned}
	\bm{S} = \bm{e}_1\bm{e}\T_1&, &
	\bm{P} = \sum_{d=1}^{2}\bm{e}_d\bm{e}\T_d&, &
	\bm{B} = \sum_{d=1}^{3}\bm{e}_d\bm{e}\T_d ,
	\end{aligned}
	\end{equation}
	where $\bm{S}$ describes the stick tensor, $\bm{P}$ the plate tensor and $\bm{B}$ the ball tensor.

	\subsubsection{Voting Process}
	%
	The main goal of Tensor Voting is to infer information represented by the tensor $\bm{K}_i$ at each position $\bm{x}_i$ by accumulating cast vote $\mathbf{V}$ from its neighborhood $\mathcal{N}$, following
	\begin{equation}
		\bm{K}_i = \sum_{\bm{x}_j \in \mathcal{N}(\bm{x}_i)} \mathbf{V}(\bm{x}_i, \bm{x}_j).
	\end{equation}
	%
	This process can be interpreted as a convolution with a predefined aligned voting field. 
	The voting fields encode the basis tensors and are derived from the 2D stick field by integration (see~\cite{Medioni2000} for more details). 
	Each input point is encoded into a tensor.
	First, if no direction is given, the tensor encodes a unit ball $\bm{B}$.
	Second, if tangents are provided, the tensor encodes a plate $\bm{P}$.
	Finally, if normals are available, the tensor encodes a stick $\bm{S}$. 
	In a case where no direction is given, a first pass of refinement is done to derive the preferred orientation information. 
	Each tensor then broadcasts each of its independent elements using an appropriate tensor field:
	\begin{equation}
	 \mathbf{V}(\bm{x}_i, \bm{x}_j) = \mathbf{V}_{\bm{S}}(\bm{x}_i, \bm{x}_j) +  \mathbf{V}_{\bm{P}}(\bm{x}_i, \bm{x}_j) +  \mathbf{V}_{\bm{B}}(\bm{x}_i, \bm{x}_j),
	\end{equation}
	where $\mathbf{V}_{\bm{S}}$ (resp. $\mathbf{V}_{\bm{P}}$ and $\mathbf{V}_{\bm{B}}$) is the vote generated by the tensor field associated to $\bm{S}$ (resp. $\bm{P}$ and $\bm{B}$).

	\subsubsection{Vote Interpretation}
	%
	The resulting generic second order symmetric tensor $\bm{K}$ is then decomposed into elementary components to extract the saliencies and the preferred direction. 
	The interpretation of these values is given in \autoref{tab:saliencies}. 
	We can then infer geometric primitives, but the procedure to extract the salient features corresponding to local maxima of the three saliency maps will not be discussed here.

	\subsubsection{k-Nearest Neighbors Closed Form Tensor Voting}
	%
	Although tensor voting is a robust technique for extracting perceptual information from point clouds, the complexity of its original formulation makes it difficult to use in robotics applications.
	We use the \ac{CFTV} formulation proposed by~\citet{Wu2016} for efficiency.
	The generic second order symmetric tensor is then computed given
	\begin{equation}\label{eq:cftv}
	\begin{aligned}%
	\bm{K}_i &= \sum_{\bm{x}_j \in \mathcal{N}(\bm{x}_i)} \mathbf{S}_{ij} & \text{with } \mathbf{S}_{ij}&=c_{ij}\mathbf{R}_{ij}\bm{K}_{j}\mathbf{R}_{ij}^{\prime} ,
	\end{aligned}%
	\end{equation}
	and
	\begin{equation}\label{eq:cftvparam}
	\begin{aligned}
	\mathbf{R}_{ij} &= \left( \mathbf{I}-2\bm{r}_{ij}\bm{r}_{ij}\T\right), &
	\mathbf{R}_{ij}^{\prime} &= \left(\mathbf{I}-\frac{1}{2}\bm{r}_{ij}\bm{r}_{ij}\T\right)\mathbf{R}\T_{ij}, \\
	\bm{r}_{ij} &= \frac{\bm{x}_i-\bm{x}_j}{\norm{\bm{x}_i-\bm{x}_j}}, &
	c_{ij} &= \exp\left(-\frac{\norm{\bm{x}_i-\bm{x}_j}^2}{\sigma} \right),
	\end{aligned}
	\end{equation}
	where $c_{ij}$ controls the strength of the vote given the distance between the two positions and the scale parameter $\sigma$; 
	$\bm{r}_{ij}$ is the normalized vector from $\bm{x}_j$ in the direction of $\bm{x}_i$; 
	and $\mathcal{N}$ is the neighborhood retrieved using an efficient \textit{k}-Nearest Neighbors (\textit{k}-NN) search (e.g., with a \textit{kD-tree}).
	As the input is generally not oriented, we still have to do a first pass by encoding $\bm{K}_j$ as a unit ball to derive a preferred direction. 
	Then, we do a second pass by encoding points with the tensors previously obtained, but with the ball component disabled~\cite{Wu2016} such as $\bm{K}_j = \left(\lambda_1 - \lambda_2 \right) \bm{S}_j + \left(\lambda_2 - \lambda_3 \right) \bm{P}_j$. 
	Once the generic tensor is computed, we decompose and interpret it as shown above.
	
	\begin{table}[t]
		\centering
		\caption{Interpretation of saliencies and preferred directions obtained by the tensor voting framework.}
		\label{tab:saliencies}
		\begin{tabularx}{\columnwidth}{Xccccc}
			\toprule
			& Geom. Primitive & Tensor & Saliency & Normals \\ 
			\midrule
			
			Surface-ness & Surface & Stick $\bm{S}$ & $\lambda_1 - \lambda_2$ & $\bm{e}_1$ \\ 
			Curve-ness & Curve & Plate $\bm{P}$ & $\lambda_2 - \lambda_3$ & $\bm{e}_1$, $\bm{e}_2$ \\ 
			Point-ness & Junction & Ball $\bm{B}$& $\lambda_3$ & $\bm{e}_1$, $\bm{e}_2$, $\bm{e}_3$ \\ 
			
			\bottomrule
		\end{tabularx}
		\vspace*{-13pt}
	\end{table}
	

\section{Derivation of density measures}\label{sec:densitymeasure}
Based on tensor voting theory, this paper presents a novel density measure for each geometric primitive.
By doing a first pass of \ac{TV} using the closed-form with an \textit{k}-NN search (\autoref{eq:cftv}), we are able to derive more information from the tensors. 
In fact we can show that $0 \le \lambda_d \le k\text{, } \forall d \in \left\lbrace 1,2,3 \right\rbrace$, where $k$ is the number of neighbors. 
As the strength of the vote through the kernel is directly dependent on the distance, we have $\lambda_d = k$ when all neighbors are at a distance $\delta=0$. 
Given this observation, the lambdas can be considered as an indicator of local density.

In the following, the $\lambda_d$ are normalized by $k$.
We can compute the expected normalized kernel strengths $\xi_D$ at a position where the density would be uniform in a $D$-hyperball of radius $\rho$ to derive the density measures.
As the strength of the vote is only dependent on the distance, and therefore only the kernel function $\mathrm{k}(\delta)=\exp\left(-\delta^2/\sigma\right)$ 
is taken into account, we compute the expectation of the kernel function given a uniform distribution $\mathcal{U}$ of the distance $\delta$ in a $D$-hyperball of radius $\rho$ with the random variables $X \sim \mathcal{U}_{\left[-1,1\right]}$ and $\mathrm{\delta}(X) \sim \rho\abs{\mathcal{U}_{\left[-1,1\right]}}^{\frac{1}{D}}$. 
We then compute the expected value of this distribution such as
\begin{equation}
	\begin{aligned} \label{eq:expectation}
		\mathbf{E}\left[\mathrm{k}(\mathrm{\delta}(X))\right] &= \int_{-\infty}^{\infty} \mathrm{pdf}_X(x) \cdot \mathrm{k}(\mathrm{\delta}(x)) \diff x \\
		&= \frac{D}{2}\left(\frac{\rho^2}{\sigma} \right)^{-\frac{D}{2}}\left(\mathrm{\varGamma}\left(\frac{D}{2}\right) - \mathrm{\varGamma}\left(\frac{D}{2}, \frac{\rho^2}{\sigma}\right) \right),
	\end{aligned}
\end{equation}
where 
$\sigma$ is the scale of the kernel vote,
$\mathrm{\varGamma}(\cdot)$ is the gamma function and $\mathrm{\varGamma}(\cdot,\cdot)$ is the incomplete gamma function. 
For $D \in \left\lbrace1,2,3\right\rbrace$, the expected kernel strengths are given by
\begin{equation}
	\begin{aligned}\label{eq:expectedvalues}
		\xi_{1} &= \frac{1}{4\rho}~\sqrt{\pi\sigma} ~\mathrm{erf}\left(\frac{\rho}{\sqrt{\sigma}}\right)\\
		\xi_{2} &= \frac{\sigma}{\rho^2}~\left(1 - \exp\left(-\frac{\rho^2}{\sigma}\right)\right) \\
		\xi_{3} & = \frac{3\sigma}{4\rho^3}~\left(\sqrt{\pi\sigma}~\mathrm{erf}\left(\frac{\rho}{\sqrt{\sigma}}\right) - 2\rho\exp\left(-\frac{\rho^2}{\sigma}\right)\right) ,
	\end{aligned}
\end{equation}
where $\mathrm{erf}(\cdot)$ is the Gauss error function, $\xi_{1}$ is the expected strength in the plate case ($D=1$), $\xi_{2}$ is the expected strength in the stick case ($D=2$) and $\xi_{3}$ is the expected strength in the ball case ($D=3$).

\begin{figure}[!t]	
	\vspace*{-8pt}
	\centering
		\includegraphics[width=0.33\columnwidth]{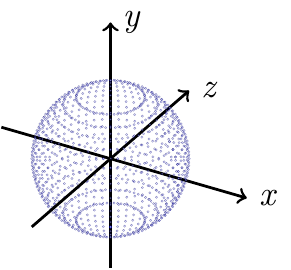}%
		\includegraphics[width=0.33\columnwidth]{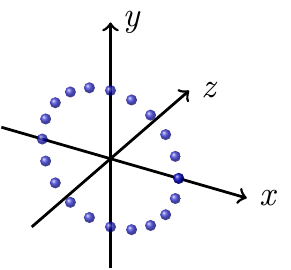}%
		\includegraphics[width=0.33\columnwidth]{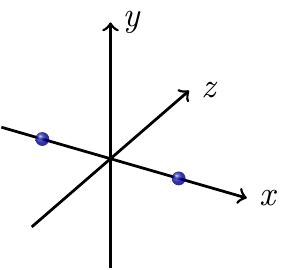}%
	\vspace*{-3pt}
	\caption{
	Ideal simplified voting situations. 
		\emph{Left:} All points are uniformly distributed on a sphere ($D=3$);  
		\emph{Middle:} All points are uniformly distributed on a circle lying in the \textit{xy} plane ($D=2$);
		\emph{Right:} All points are uniformly distributed on the extremities of a segment along the \textit{x} axis ($D=1$).
	}
	\label{fig:idealvote}
	\vspace*{-13pt}
\end{figure}

The associated eigenvalues $\hat{\lambda}_d$ can be derived if we consider, for each component, the ideal cases illustrated by \autoref{fig:idealvote} where each voter strength is $\xi_D$. 
By developing \autoref{eq:cftv} with $c_{ij}=\xi_D$ and taking $r_{ij}$ as the integral variable on the considered domain (i.e., all possible orientations in $D$ dimensions), we deduce for each case the expected eigenvalues $\hat{\lambda}_d$.
%
%
%
We are now able to interpret the saliencies obtained by the closed-form \ac{TV} where $\bm{K}_j = \bm{I}, \forall j$ as a measure of local density. 
We can therefore compare the values with the expected saliencies (summarized in \autoref{tab:expectedsaliencies}) to control the density of each geometric primitive.

\section{Spectral Decomposition Filter (SpDF): Overview} \label{sec:spdf}
%
The method presented in this article aims to reduce the number of points while preserving as much as possible the topology of the point cloud using geometric primitives (i.e., curves, surface, and junction). Note that it is not limited to plane, line and point as the tensor voting framework allows to detect more generic geometric primitives.
A major challenge in robotics applications is the non-uniformity of scans acquired with \ac{LIDAR} sensors. 
In fact most of sampling algorithms are designed for uniform point clouds. 
This problematic is addressed by proposing a new efficient method to make the density uniform for each of the three geometric primitives we consider.
Our method can be divided into three main steps: 1) making the density uniform for each geometric primitive; 2) rejecting outliers according to the confidence of the geometric information; and 3) subsampling.
	
	\subsection{Making the density uniform}	
\begin{figure*}[htbp]
		\centering
		\begin{subfigure}[]{0.50\textwidth}
			\hspace*{-30pt}
			\includegraphics[width=1.15\textwidth]{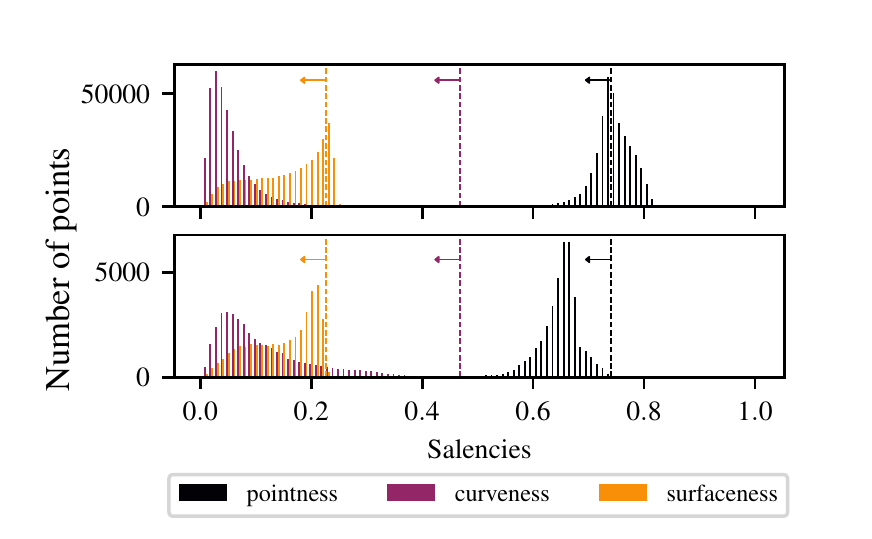}%
		\end{subfigure}%
		\begin{subfigure}[]{0.40\textwidth}
			\begin{subfigure}[]{\textwidth}%
				\includegraphics[width=1.05\columnwidth]{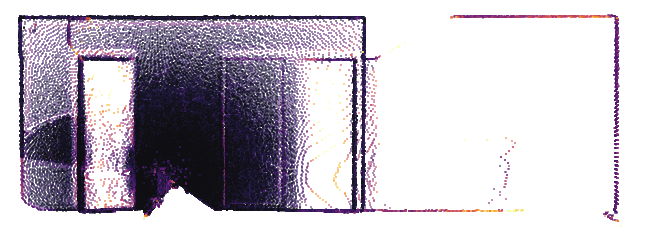}
			\end{subfigure}
			\begin{subfigure}[]{\textwidth}%
				\includegraphics[width=\columnwidth]{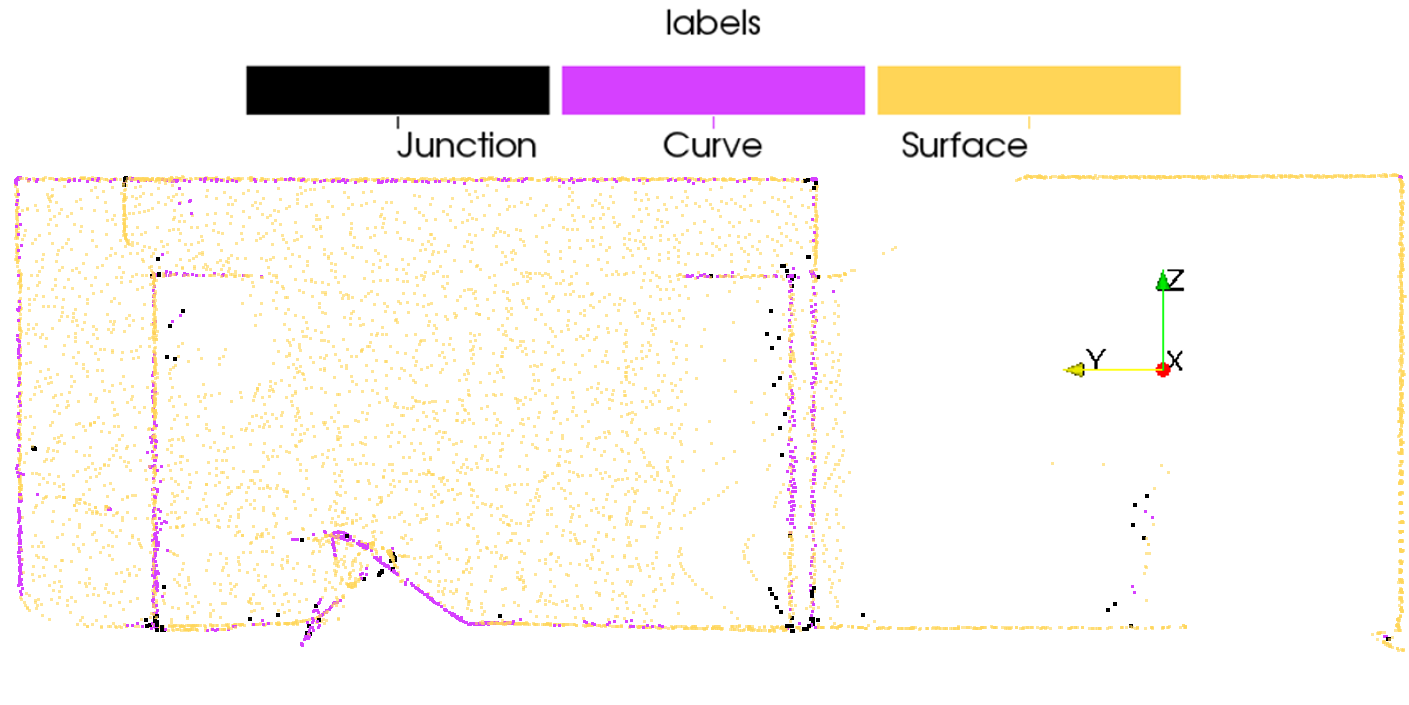}
			\end{subfigure}%
		\end{subfigure}%
		\vspace*{-10pt}	
		%
		\caption{
			Illustration of our method to reduce and make uniform a point cloud from \num{370}\si{\kilo} to \num{40}\si{\kilo} points.
			\emph{Left}: Convergence of the saliencies below their expected values (represented by the vertical dashed lines) implying a uniform density on each geometric primitive (with $\sigma = \rho = 0.2$). 
				\emph{Top-Left:} The histogram of the initial saliencies distribution; 
				\emph{Bottom-Left:} The histogram of the resulting distribution after making the density uniform.
			\emph{Top-Right}: The original point cloud with uneven density.
			\emph{Bottom-Right}: The resulting point cloud augmented with labels obtained by a second pass of tensor voting. Density is uniform, and geometric primitives are clearly identified.
		}
		\label{fig:method}
		\vspace*{-10pt}
	\end{figure*}

	%
	Using the new local density measure on each geometric primitive, the point cloud can be made uniform as follows.
	An iterative procedure allows to progressively decimate primitives where the saliency measures are higher than the expected values. 
	The saliencies are recomputed using a pass of \ac{TV} with tensors encoded as unit balls.
	The algorithm stops when the number of points is stable, which means that the saliencies distributions have converged below the expected values, as shown by \autoref{fig:method}-\emph{Left}.
	Therefore, the densities are uniform around each primitive allowing us to detect them more clearly.
	Otherwise, most dense areas will be detected as junction because noise will predominated.
	An example of the result of making the density uniform is given by \autoref{fig:method}-\emph{Right}.
	%
	
	\begin{table}[!t]
		\centering
		\caption{
			Expected eigenvalues and saliencies in the case of a uniform density in a $D$-hyperball. 
		}
		\label{tab:expectedsaliencies}
		\begin{tabularx}{\columnwidth}{ccXc}
			\toprule
			& D & Eigenvalues & Saliency\\
			\midrule
			
			Curve-ness ($\bm{P}$) & 1 & $\hat{\lambda}_1 = \hat{\lambda}_2 = \xi_{1} \text{ and } \hat{\lambda}_3 =  \frac{1}{2}~\xi_{1}$ &$\frac{1}{2}~\xi_{1}$  \\
			Surface-ness ($\bm{S}$) & 2 & $\hat{\lambda}_1 = \xi_{2}\text{ and }\hat{\lambda}_2 = \hat{\lambda}_3 = \frac{3}{4}~\xi_{2}$ & $\frac{1}{4}~\xi_{2}$  \\
			Point-ness ($\bm{B}$) & 3 & $\hat{\lambda}_1 = \hat{\lambda}_2 = \hat{\lambda}_3 = \frac{5}{6}~\xi_{3}$ & $\frac{5}{6}~\xi_{3}$  \\
			
			\bottomrule
		\end{tabularx}
		\vspace*{-13pt}
	\end{table}
		
	\subsection{Rejecting outliers}
	%
	Given the saliencies computed with a last pass of \ac{TV} with the ball component disabled, each point is then labeled into \textit{junction}, \textit{curve} or \textit{surface}, and the saliency associated (respectively point-ness, curve-ness or surface-ness) encodes the confidence in this labeling. 
	It provides a high level description in terms of geometry as shown by \autoref{fig:method}-\emph{Bottom-Right}.
	Points with a confidence higher than $t$~\% of the maximum confidence of the considered geometric primitive are kept (e.g., in our experiments we use $t=10\%$). This heuristic allows to reject outliers having a low confidence in their measure.
	
	\subsection{Sampling on geometric primitives}
	%
	At this step, the point cloud is uniform, outliers have been rejected, and each point is labeled.
	Given the parameters $\sigma$ and $\rho$, the point cloud has already been considerably reduced.
	Indeed, high density areas weigh for a large amount of points.
	In order to reduce even more the number of points, several strategies can be designed.
	%
		
		One can tune $\sigma$, the scale of vote (\autoref{eq:cftvparam}), and $\rho$, the radius of uniformity (\autoref{eq:expectedvalues}), to reduce the number of points.
		Since $\sigma$ is linked to the vote process of tensor voting, it should not be used with this intention.
		One the other hand, $\rho$ directly controls the number of points within the $D$-hyperball, keeping the number of neighbors $k$ constant.
		%
		
		One would take into account the spatial distribution.
		To ensure the conservation of the topology, this paper proposes considering a spatial sampling on each geometric primitive.
		Concretely, the point cloud is divided into three sub-point clouds where each sub-point cloud represents only one geometric primitive. 
		Spatial segmentation based on an \textit{octree} is then conducted on each of them to sub-sample using the centroid method given a number $n$ of points in each cells.
		The ratio \textit{surface-plane-junction} is kept during this process and the three sub-point clouds are reassembled at the end. 	
	

\section{Experimental Setup}\label{sec:setupexp}

To evaluate the impact of the number of points on the registration process, several methods from the state-of-the-art had been implemented in the open-source modular library for \ac{ICP} named \texttt{libpointmatcher}, introduced in~\cite{Pomerleau2013} and available online\footnote{ \href{https://github.com/ethz-asl/libpointmatcher}{https://github.com/ethz-asl/libpointmatcher}}.
The nine evaluated filters are resumed in \autoref{tab:results}.
For more clarity and  without loss of generality, only eight of them will be presented for all environments concatenated in \autoref{fig:results}.

Working under the hypothesis of robotics applications, this paper presents an in-depth comparison of sampling algorithms on 1) structured (with \num{45} pairs of scans), 2) semi-structured (with \num{32} pairs of scans), and 3) unstructured point clouds (with \num{32} pairs of scans), using the datasets~\citetitle{Pomerleau2012b}~\cite{Pomerleau2012b}.

To evaluate the accuracy of the registration, we calculate separately the translation error part, $\varepsilon_t$, and the rotational error part, $\varepsilon_r$, the same way it is done in~\citetitle{Pomerleau2013}~\cite{Pomerleau2013}.
	%
For each method, the range of parameters influencing the number of points to obtain between \num{1000} and \num{2e5} points was determined and resumed in \autoref{tab:results}.
We performed \num{2500} registration using \ac{ICP} (\num{5000} for our baseline \texttt{Random}) across the range for each method on each pair of scans for each dataset.
	
As \ac{ICP} needs a prior for fine registration (i.e., the initial transformation $\check{\bm{T}}$) to compute the transformation between two point clouds, we applied a uniform perturbation on the ground truth transformation, such that $\check{\bm{T}} = \exp(\bm{\varsigma})\bm{T}_{gt}$, with $\bm{\varsigma} \in \mathfrak{se}(3)$. 
For our experiments, a perturbation sampled from a uniform distribution of \SI{50}{cm} was applied on the translation, and from a uniform distribution of \SI{20}{\degree} on the rotation.
%
During the data filtering step, we applied the filter on both the reading and the reference. 
The data association is conducted by matching the two closest neighbors.
We rejected the outliers according to a trimmed distance.
We limited the scope our experiments to a \textit{point-to-plane} version of \ac{ICP}, as it tends to perform better in those datasets~\cite{Pomerleau2013}, and we keep the evaluation of different error metrics for future works.

\begin{table*}[htbp]
	\centering
	\caption{
		Comparison of the impact of the number of points on the registration process. For each method, the range of parameters and its signification are given. Median errors in translation and rotation are reported for each type of environment.
	}
	\label{tab:results}
	\begin{threeparttable}
	\begin{tabularx}{\textwidth}{@{}llXcccc@{}}	
		\toprule
		\multirow{2}{*}{Method} & \multirow{2}{*}{Parameter description} & \multirow{2}{*}{Range\tnote{1}} & \multicolumn{4}{>{\centering\arraybackslash}c}{Median errors $\left(\varepsilon_t\text{ [\si{\meter}]};~\varepsilon_r\text{ [\si{deg}]}\right)$} \\
		\cmidrule(l){4-7}
		{} & {} & {} & Structured & Semi-structured & Unstructured & \textbf{All} \\
		\midrule
		\texttt{Random} (baseline) & prob. to keep point & $\left[0.004~;~1.\right]$ & $\left(0.110;~2.001\right)$ & $\left(0.011;~0.291\right)$ & $\left(0.330;~3.671\right)$ & $\left(0.066;~1.102\right)$ \\
		\texttt{Max Density}\tnote{2} & nb. max of points by \si{\meter\cubed} & $\left[16.8~;~506\text{\si{\kilo}}\right]$ & $\left(0.018;~\textbf{0.489}\right)$ & $\left(0.010;~0.241\right)$ & $\left(\textbf{0.019};~\textbf{0.213}\right)$ & $\left(\textbf{0.016};~\textbf{0.297}\right)$ \\
		\texttt{SSNormal}\tnote{2} & nb. of neighbors to merge & $\left[253~;~3.\right]$ & $\left(0.088;~1.847\right)$ & $\left(0.016;~0.340\right)$ & $\left(0.313;~2.605\right)$ & $\left(0.065;~1.273\right)$ \\
		\texttt{Octree (centroid)}& nb. max of points by cell & $\left[1000~;~1.\right]$ & $\left(0.046;~1.025\right)$ & $\left(0.008;~0.256\right)$ & $\left(0.113;~1.134\right)$ & $\left(0.019;~0.402\right)$ \\
		\texttt{Voxel} & size max of the cell & $\left[2.49~;~0.001\right]$ & $\left(\textbf{0.010};~\textbf{0.452}\right)$ & $\left(0.008;~0.232\right)$ & $\left(\textbf{0.018};~\textbf{0.219}\right)$ & $\left(\textbf{0.012};~\textbf{0.265}\right)$ \\
		\texttt{NSS}~\cite{Rusinkiewicz2001} & nb. of points to keep & $\left[1000~;~n\right]$ & $\left(0.372;~9.895\right)$ & $\left(0.013;~0.293\right)$ & $\left(0.257;~3.419\right)$ & $\left(0.201;~3.229\right)$ \\
		\texttt{CovS}~\cite{Gelfand2003} & nb. of points to keep & $\left[1000~;~n\right]$ & $\left(0.340;~6.398\right)$ & $\left(0.019;~0.363\right)$ & $\left(0.327;~3.144\right)$ & $\left(0.173;~1.973\right)$ \\
		\texttt{Spatial \ac{SpDF}} (ours) & nb. of points to keep & $\left[1000~;~n\right]$ & $\left(\textbf{0.013};~\textbf{0.497}\right)$ & $\left(0.008;~0.237\right)$ & $\left(\textbf{0.020};~\textbf{0.242}\right)$ & $\left(\textbf{0.013};~\textbf{0.314}\right)$ \\
		\texttt{\ac{SpDF}} (ours) & radius of uniformity & $\left[1.35~;~0.1\right]$ & $\left(0.020;~0.599\right)$ & $\left(0.012;~0.256\right)$ & $\left(\textbf{0.025};~\textbf{0.253}\right)$ & $\left(\textbf{0.018};~\textbf{0.278}\right)$ \\
		\bottomrule
	\end{tabularx}
	\begin{tablenotes}
		\item[1] The range is given as $\left[a;~b\right]$, where $a$ gives the smallest number of points and $b$ preserves all the points ($n$ being the total number of points).
		\item[2] Methods from the \texttt{libpointmatcher}, one working on density (\texttt{MaxDensity}) and the other grouping points on surfaces (\texttt{SSNormal})
	\end{tablenotes}
	\end{threeparttable}
	\vspace*{-10pt}
\end{table*}

\section{Results \& Discussion}\label{sec:results}

This paper presents a quantitative experiment of the influence of the number of points on the registration accuracy.
It also highlights the importance of having a uniform density to be able to preserve the topology of the point cloud on real world dataset.

	\subsubsection{Comparison on registration accuracy}
\begin{figure}[htbp]
		\centering
		\vspace*{-7pt}
		\hspace*{-8pt}
		\includegraphics[width=1.05\columnwidth]{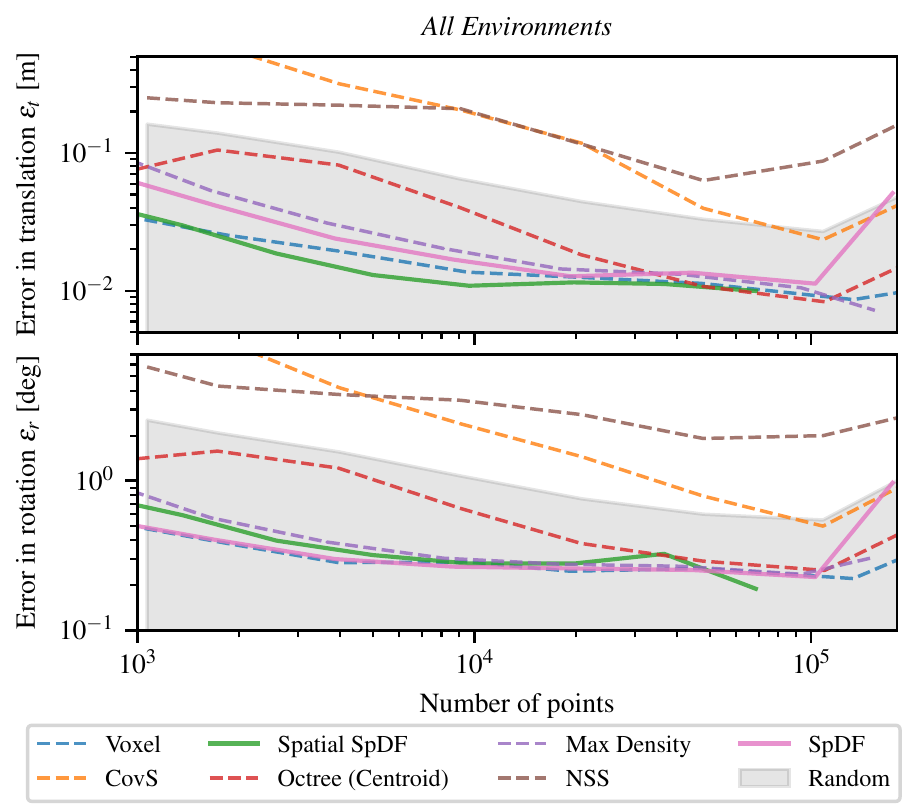}
		\caption{
			Influence of the number of points on the registration process. 
			The gray area represents the errors inferior to the baseline. 
			Our methods are displayed in solid-line.
			\emph{Top}: error in translation $\varepsilon_t$ in \si{\meter};
			\emph{Bottom}: error in rotation $\varepsilon_r$ in \si{deg}.
			Both translation and rotational errors show the same patterns. 
			%
		}
		\vspace*{-23pt}
		\label{fig:results}
	\end{figure}

	\autoref{fig:results} presents the translation (\emph{Top}) and the rotational (\emph{Bottom}) errors as functions of the number of points in the point clouds for all environments concatenated.
	The gray area represents the errors inferior to our baseline (\texttt{Random}), the solid-lines correspond to our methods and the dashed-lines to the other methods. 
	Details for each type of environment are reported in \autoref{tab:results}.
	Each environment shows its own variation of errors, but the methods behave similarly.
	The semi-structured environment one does not give a lot of information as all methods perform well within centimeters for the translation error and under \SI{0.5}{deg} for the rotational error.
	Unsurprisingly, errors are greater for the unstructured than the structured environment, and the differences between methods are better highlighted.
	
	Both translation ($\varepsilon_t$) and rotational ($\varepsilon_r$) errors show the same patterns. 
	Each method presents a bump around \num{e5} points. 
	With more points the dense areas over-constrained the minimization process.
	With less points, the minimization process is under-constrained.
	Both situations lead to less accurate results.	 
	The evaluated methods are significantly more accurate than our baseline except
	both \texttt{\ac{NSS}} and \texttt{\ac{CovS}}, which perform worse than \texttt{Random}, with an error of \SI{20}{\centi\meter} against \SI{7}{\centi\meter} for the translation.
	They are unable to manage uneven density and large-scale point cloud performing poorly for all types of environment.
	These algorithms need to be adapted for an application in the context of robotics.

	In particular, \texttt{Octree} starts diverging quite sooner than the others, around \num{3e4} points when the number of points decreases.
	Indeed, spatial segmentation like \texttt{Octree} performs well in a certain measure as it is able to preserve the spatial distribution for a large number of points but suffers from the uneven density distribution for a small number of points.
	At \num{2e3} points, the error of \texttt{Octree} decreases due to the concatenation of the environments. Indeed, the error is stagnating for the unstructured one.
	Eventually, the \texttt{Voxel} version performs better as it preserves the spatial distribution whatever the density.
	It however does not differentiate the geometric primitives within the cells, then losing these information.
	
	Using only the density, \texttt{MaxDensity} leads to more accurate alignments than the baseline showing a translation error under \SI{2}{\centi\meter} and a rotational error less than \SI{0.5}{deg}. 
	However, the densities calculation excludes the local topology and use a spherical approximation to compute it.
	%
	Our methods (\texttt{SpDF} and Spatial \texttt{SpDF}), along \texttt{Voxel}, show the best results from all the evaluated methods, with an error in translation of \SI{1}{\centi\meter} and an error in rotation of \SI{0.3}{deg}.
	Eventually, our methods inherits the qualities of the previous methods and manage efficiently large point clouds with uneven densities by maintaining a given density for each geometric element.
	By sampling spatially on each geometric feature we further preserve the details and the topology of the point cloud.

	\subsubsection{Uniform density on real-world large-scale dataset}
		
	\autoref{fig:method} illustrates that our methods are able to make the density uniform on each geometric primitive.
	We evaluated the proposed method against the baseline random sampling on a real-world dataset. 
	\autoref{fig:laval} shows the qualitative result of a sampled point cloud from \num{4.65} million points to \num{30}\si{\kilo} points.
	In particular, from the results for the random sampling method (\emph{Left}), we can see that only dense areas have been kept and many geometric details have been deleted. 
	Contrarily, our method's results (\emph{Right}) show that most of the details have been preserved and the density is more uniform.
	
\section{Conclusion}\label{sec:conclusion}
%
This paper presents a novel sampling algorithm aiming at better supporting the \ac{ICP} algorithm. 
This method build on spectral decompositions applied to point clouds in order to obtain a density better suited for \ac{ICP}.
Moreover, this sampling algorithm works on each geometric primitive separately in order to reject outliers and subsample points.

We validated these observations through quantitative and qualitative results.
%
Our methods perform successfully on large-scale maps, where the density is non-uniform. 
We proposed a solution to the limitations of Tensor Voting by deriving a new measure of density directly from the saliencies.
This measure allows to preserve the topology of the point cloud by maintaining the geometric primitives.
%
By taking into account the spatial distribution of each geometric primitive, we efficiently subsample the point cloud, thus reducing the number of points without losing accuracy during the registration process. 

We provide a second order symmetric tensor representation for each point (i.e., a Gaussian representation).
Future work will be done to incorporate this information directly in the minimization process of \ac{ICP}, inspired by~\cite{Segal2009,Stoyanov2012}.

\section*{Acknowledgment}

This work was partially supported by the French program WOW! Wide Open to the World in the context of the project I-SITE Clermont CAP 20-25.

\printbibliography

\end{document}